\documentclass[journal,a4paper]{article}
\usepackage[margin=0.9in]{geometry} 
\usepackage{setspace}
\usepackage{amsmath,amsfonts}
\usepackage{algorithmic}
\usepackage{array}
\usepackage[caption=false,font=normalsize,labelfont=sf,textfont=sf]{subfig}
\usepackage{textcomp}
\usepackage{stfloats}
\usepackage{url}
\usepackage{verbatim}
\usepackage{booktabs}
\usepackage{graphicx}
\usepackage{multirow}
\usepackage{acro} 
\usepackage[switch]{lineno} 
\usepackage{authblk}
\DeclareAcronym{USA}{
  short = U.S.,
  long  = United States
}

\DeclareAcronym{UK}{
  short = UK,
  long  = United Kingdom
}

\DeclareAcronym{LTHT}{
  short = LTHT,
  long  = Leeds Teaching Hospital Trust
}

\DeclareAcronym{2D}{
  short = 2D,
  long  = Two-dimensional
}

\DeclareAcronym{3D}{
  short = 3D,
  long  = Three-dimensional
}

\DeclareAcronym{HIFU}{
  short = HIFU,
  long  = High-Intensity Focused Ultrasound
}

\DeclareAcronym{FUS}{
  short = FUS,
  long  = Focused Ultrasound
}

\DeclareAcronym{US}{
  short = US,
  long  = Ultrasound
}

\DeclareAcronym{LUS}{
  short = LUS,
  long  = Lung Ultrasound
}

\DeclareAcronym{MRI}{
  short = MRI,
  long  = Magnetic Resonance Imaging
}

\DeclareAcronym{MRT}{
  short = MRT,
  long  = Magnetic Resonance Thermometry
}

\DeclareAcronym{CT}{
  short = CT,
  long  = Computed Tomography
}

\DeclareAcronym{CBCT}{
  short = CT,
  long  = Cone Beam Computed Tomography
}

\DeclareAcronym{B-mode}{
  short = B-mode,
  long  = Brightness Mode
}

\DeclareAcronym{QA}{
  short = QA,
  long  = Quality Assurance
}

\DeclareAcronym{IEC}{
  short = IEC,
  long  = International Electrotechnical Commission
}

\DeclareAcronym{MI}{
  short = MI,
  long  = Mechanical Index
}

\DeclareAcronym{TI}{
  short = TI,
  long  = Thermal Index
}

\DeclareAcronym{TGC}{
  short = TGC,
  long  = Time Gain Compensation
}

\DeclareAcronym{ROI}{
  short = ROI,
  long  = Region of Interest
}

\DeclareAcronym{FOV}{
  short = FOV,
  long  = field of view
}

\DeclareAcronym{BLAS}{
  short = BLAS,
  long  = B-line Artefact Score
}

\DeclareAcronym{RADEX}{
  short = RADEX,
  long  = Radiology Data Extraction Tool
}

\DeclareAcronym{US-QA-3D}{
  short = US-QA-3D,
  long  = Ultrasound Quality Assurance in 3-Dimensions
}

\DeclareAcronym{GPU}{
  short = GPU,
  long  = Graphics Processing Unit
}

\DeclareAcronym{CPU}{
  short = CPU,
  long  = Central Processing Unit
}

\DeclareAcronym{RAM}{
  short = RAM,
  long  = Random Access Memory
}

\DeclareAcronym{HDMI}{
  short = HDMI,
  long  = High-Definition Multimedia Interface
}
\DeclareAcronym{USB}{
  short = USB,
  long  = Universal Serial Bus
}

\DeclareAcronym{VIA}{
  short = VIA,
  long  = VGG Image Annotator
}

\DeclareAcronym{UMLS}{
  short = UMLS,
  long  = Unified Medical Language System 
}

\DeclareAcronym{regex}{
  short = Regex,
  long  = Regular expressions 
}

\DeclareAcronym{GUI}{
  short = GUI,
  long  = Graphical User Interface 
}

\DeclareAcronym{CAD}{
  short = CAD,
  long  = Computer-Aided Design 
}

\DeclareAcronym{MRML}{
  short = MRML,
  long  = Medical Reality Modelling Language 
}

\DeclareAcronym{VTK}{
  short = VTK,
  long  = Visualisation Toolkit 
}

\DeclareAcronym{ICP}{
  short = ICP,
  long  = Iterative Closest Point 
}

\DeclareAcronym{MPR}{
  short = MPR,
  long  = multiplanar reconstruction
}

\DeclareAcronym{API}{
  short = API,
  long  = application programming interface
}

\DeclareAcronym{RUS}{
  short = RUS,
  long  = Robotic ultrasound
}

\DeclareAcronym{DoF}{
  short = DoF,
  long  = Degrees of Freedom
}

\DeclareAcronym{EM}{
  short = EM,
  long  = electromagnetic
}

\DeclareAcronym{EMI}{
  short = EMI,
  long  = electromagnetic interference
}

\DeclareAcronym{LDLJ}{
  short = LDLJ,
  long  = log dimensionless jerk
}

\DeclareAcronym{DH}{
  short = DH,
  long  = Denavit-Hartenberg
}

\DeclareAcronym{SPARC}{
  short = SPARC,
  long  = spectral arc length
}

\DeclareAcronym{AI}{
  short = AI,
  long  = Artificial Intelligence
}

\DeclareAcronym{ML}{
  short = ML,
  long  = Machine Learning
}

\DeclareAcronym{DL}{
  short = DL,
  long  = Deep Learning
}

\DeclareAcronym{CNNs}{
  short = CNNs,
  long  = Convolutional Neural Networks
}

\DeclareAcronym{CNN}{
  short = CNN,
  long  = Convolutional Neural Network
}

\DeclareAcronym{grad-CAM}{
  short = grad-CAM,
  long  = gradient-weighted Class Activation Maps
}

\DeclareAcronym{NLP}{
  short = NLP,
  long  = Natural Language processing
}

\DeclareAcronym{LLMs}{
  short = LLMs,
  long  = Large Language Models
}

\DeclareAcronym{BERT}{
  short = BERT,
  long  = Bidirectional Encoder Representations from Transformers
}

\DeclareAcronym{AR}{
  short = AR,
  long  = Augmented Reality
}

\DeclareAcronym{DSC}{
  short = DSC,
  long  = Dice Similarity Coefficient
}

\DeclareAcronym{FPS}{
  short = FPS,
  long  = frames per second
}

\DeclareAcronym{SD}{
  short = SD,
  long  = standard deviation
}

\DeclareAcronym{HD}{
  short = HD,
  long  = Hausdorff Distance
}

\DeclareAcronym{HD95}{
  short = HD95,
  long  = 95th Percentile Hausdorff Distance
}

\DeclareAcronym{RMS}{
  short = RMS,
  long  = Root mean square
}

\DeclareAcronym{FDA}{
  short = FDA,
  long  = Food and Drug Administration
}

\DeclareAcronym{MNG}{
  short = MNG,
  long  = multinodular goitre
}

\DeclareAcronym{PACS}{
  short = PACS,
  long  = Patient Archive and Communication System
}

\DeclareAcronym{CRIS}{
  short = CRIS,
  long  = Computerised Radiology Information System
}

\DeclareAcronym{BTA}{
  short = BTA,
  long  = British Thyroid Association
}

\DeclareAcronym{ETA}{
  short = ETA,
  long  = European Thyroid Association
}

\DeclareAcronym{ACR}{
  short = ACR,
  long  = American College of Radiologists
}

\DeclareAcronym{TIRADS}{
  short = TIRADS,
  long  = Thyroid Imaging Reporting and Data System
}

\DeclareAcronym{IFNgamma}{
  short = IFN$\gamma$,
  long  = interferon gamma
}

\DeclareAcronym{APCs}{
  short = APCs,
  long  = Antigen-presenting cells
}

\DeclareAcronym{DAMPs}{
  short = DAMPs,
  long  = Damage-associated molecular patterns
}

\DeclareAcronym{HMGB1}{
  short = HMGB1,
  long  = High mobility group box 1 
}

\DeclareAcronym{CRT}{
  short = CRT,
  long  = calreticulin
}

\DeclareAcronym{HSPs}{
  short = HSPs,
  long  = heatshock proteins
}

\DeclareAcronym{RAGE}{
  short = RAGE,
  long  = Receptor for advanced glycation end-products
}

\DeclareAcronym{DCs}{
  short = DCs,
  long  = dendritic cells
}

\DeclareAcronym{HSP70}{
  short = HSP70,
  long  = 70 kilodalton heat shock proteins
}

\DeclareAcronym{CEA}{
  short = CEA,
  long  = carcinoembryonic antigen
}

\usepackage{fancyhdr}
\usepackage{tikz}

\newcommand\submittedtext{%
  \footnotesize This work has been submitted to the IEEE for possible publication. Copyright may be transferred without notice, after which this version may no longer be accessible.}

\newcommand\submittednotice{%
\begin{tikzpicture}[remember picture,overlay]
\node[anchor=south,yshift=10pt] at (current page.south) {\fbox{\parbox{\dimexpr0.65\textwidth-\fboxsep-\fboxrule\relax}{\submittedtext}}};
\end{tikzpicture}%
}

\begin{document}
\title{A Unified Platform and Quality Assurance Framework for 3D Ultrasound Reconstruction with Robotic, Optical, and Electromagnetic Tracking}
\author[1,2]{Lewis Howell}
\author[1,2]{Manisha Waterston}
\author[3,4]{Tze Min Wah}
\author[2]{James H. Chandler}
\author[2,4]{James R. McLaughlan}

\affil[1]{School of Computer Science, University of Leeds, Leeds, LS2 9JT}
\affil[2]{School of Electronic and Electrical Engineering, University of Leeds, Leeds, LS2 9JT}
\affil[3]{St James’s University Hospital, Leeds Teaching Hospitals NHS Trust, LS9 7TF}
\affil[4]{Leeds Institute of Medical Research, University of Leeds, St James' University Hospital, Leeds, LS9 7TF}  


\maketitle
\thispagestyle{fancy}

\begin{abstract}

\ac{3D} \ac{US} can facilitate diagnosis, treatment planning, and image-guided therapy. However, current studies rarely provide a comprehensive evaluation of volumetric accuracy and reproducibility, highlighting the need for robust \ac{QA} frameworks, particularly for tracked \ac{3D} \ac{US} reconstruction using freehand or robotic acquisition.

This study presents a \ac{QA} framework for \ac{3D} \ac{US} reconstruction and a flexible open source platform for tracked \ac{US} research. A custom phantom containing geometric inclusions with varying symmetry properties enables straightforward evaluation of optical, electromagnetic, and robotic kinematic tracking for \ac{3D} \ac{US} at different scanning speeds and insonation angles. 

A standardised pipeline performs real-time segmentation and \ac{3D} reconstruction of geometric targets (DSC = 0.97, FPS = 46) without GPU acceleration, followed by automated registration and comparison with ground-truth geometries. Applying this framework showed that our robotic \ac{3D} \ac{US} achieves state-of-the-art reconstruction performance (DSC-3D = 0.94~$\pm$~0.01, HD95 = 1.17~$\pm$~0.12), approaching the spatial resolution limit imposed by the transducer.

This work establishes a flexible experimental platform and a reproducible validation methodology for \ac{3D} \ac{US} reconstruction. The proposed framework enables robust cross-platform comparisons and improved reporting practices, supporting the safe and effective clinical translation of \ac{3D} ultrasound in diagnostic and image-guided therapy applications.

\end{abstract}

\vspace{0.5in}
\submittednotice

\clearpage

\section{Introduction}

Ultrasound (US) is widely used to image soft tissues, providing a non-ionising, low-cost, portable, and real-time imaging modality~\cite{chenhao_ultrasound_2026}. Beyond diagnostic applications, \ac{US} also supports intra-operative guidance~\cite{bekedam_intra-operative_2021}, image-guided biopsy~\cite{chatelain_3d_2015}, microwave and radiofrequency ablation~\cite{li_multimodality_2018, boers_3d_2024}, and emerging non-invasive modalities such as \ac{HIFU} and histotripsy~\cite{howell_histotripsy_2025, an_ultrasound_2017, ng_early_2026}. In these applications, spatial understanding of the relative shapes and positions of anatomical structures is critical to ensure diagnostic accuracy, safety, and treatment efficacy. However, traditional \ac{2D} \ac{US} provides a limited cross-sectional field of view, which reduces options for volumetric analysis and makes \ac{US}-guided applications dependent on the operator's scanning technique and skill in geometric reasoning~\cite{huang_review_2017, wang_task_2022}. 

Tracked Three-dimensional (\ac{3D}) \ac{US} addresses this limitation by combining conventional \ac{2D} imaging with position and orientation tracking to reconstruct \ac{3D} anatomy~\cite{mohamed_survey_2019, peng_recent_2022}. Compared with dedicated \ac{3D} transducers, this approach enables reconstruction over larger areas, uses widely available and more cost-effective \ac{2D} transducers and machines, and can be readily integrated with imaging-guidance systems ~\cite{mozaffari_freehand_2017, mohamed_survey_2019, chung_freehand_2017}. To acquire position and orientation tracking data, optical and \ac{EM} tracking have been used in freehand approaches~\cite{peng_recent_2022}. For example, Chung~\textit{et al.} demonstrated freehand optical-tracked \ac{US} for carotid artery \ac{3D} imaging~\cite{chung_freehand_2017}, while Krönke~\textit{et al.} applied \ac{EM}-tracked \ac{US} for thyroid volumetry, demonstrating improved repeatability relative to standard \ac{2D} approaches~\cite{kronke_tracked_2022}. More recently, robotic \ac{US} has improved scanning stability, trajectory control, and pose estimation~\cite{priester_robotic_2013}. Faoro \textit{et al.} used robotic \ac{US} for \ac{3D} vessel reconstruction as a pre-treatment planning step for transcarotid revascularisation~\cite{faoro_artificial_2023}, Sun \textit{et al.} demonstrated \ac{3D} musculoskeletal imaging~\cite{sun_automatic_2025}, and Zhou \textit{et al.} proposed a robotic \ac{US} system for kidney and lesion imaging for \ac{HIFU} imaging-guidance. Robotic kinematic tracking can be used independently for pose estimation or combined with optical or \ac{EM} tracking to simplify co-registration with patient or tool coordinate frames, supporting multimodal image guidance and interventional workflows~\cite{schmidt_tracking_2024}. 

Despite widespread investigation of tracked \ac{US} for \ac{3D} reconstruction, there is currently no standardised Quality Assurance (\ac{QA}) framework specifically designed to validate tracked \ac{3D} \ac{US} reconstruction systems. Existing commercial \ac{QA} phantoms are expensive and intended primarily to assess \ac{2D} or dedicated \ac{3D} transducers~\cite{mozaffari_freehand_2017}. Consequently, most existing tracked \ac{US} studies report task-specific outcomes, such as \ac{3D} measurement accuracy for a specific organ, but lack a comprehensive evaluation of their systems' capabilities. Task-specific metrics are often also dependent on segmentation of the test object, meaning that errors associated with calibration, tracking, and \ac{3D} reconstruction cannot be isolated from those of anatomical segmentation models (often black-box \ac{AI}-based) used to process the \ac{US} images. The robustness of these systems is also rarely assessed, particularly how scanning speed and insonation angle affect the quality and reliability of \ac{3D} reconstruction. This makes cross-system comparisons difficult, thereby limiting evidence-based development.

To address this, we present two complementary contributions. First, we introduce a flexible, open source tracked \ac{US} platform for \ac{3D} reconstruction that supports interchangeable pose estimation using kinematic, optical, or \ac{EM} tracking within a unified coordinate system. Second, we present an open source \ac{QA} framework for evaluating tracked \ac{3D} \ac{US} reconstruction, including a purpose-designed agar phantom with geometric inclusions (sphere, ellipsoid, cylinder, and triangular prism) to enable direct comparison with ground-truth volumes. Using controlled scanning and simultaneous multimodal tracking, we compare kinematic, optical, and \ac{EM} tracking for \ac{3D} reconstruction, providing insights for future research.

\section{Methods}
This study presents a flexible platform for \ac{3D} \ac{US} that integrates conventional brightness-mode (B-mode) \ac{US} imaging with robotic, optical, or \ac{EM} tracking. The workflow for \ac{3D} reconstruction comprised four stages: (A) tracked image acquisition, (B) calibration and spatial registration, (C) segmentation and \ac{3D} reconstruction, (D) quality assurance.

\subsection{Tracked Image Acquisition}
\label{sec:methods:acquisition}
\begin{figure*}[htbp]
\centering
\includegraphics[width=\textwidth]{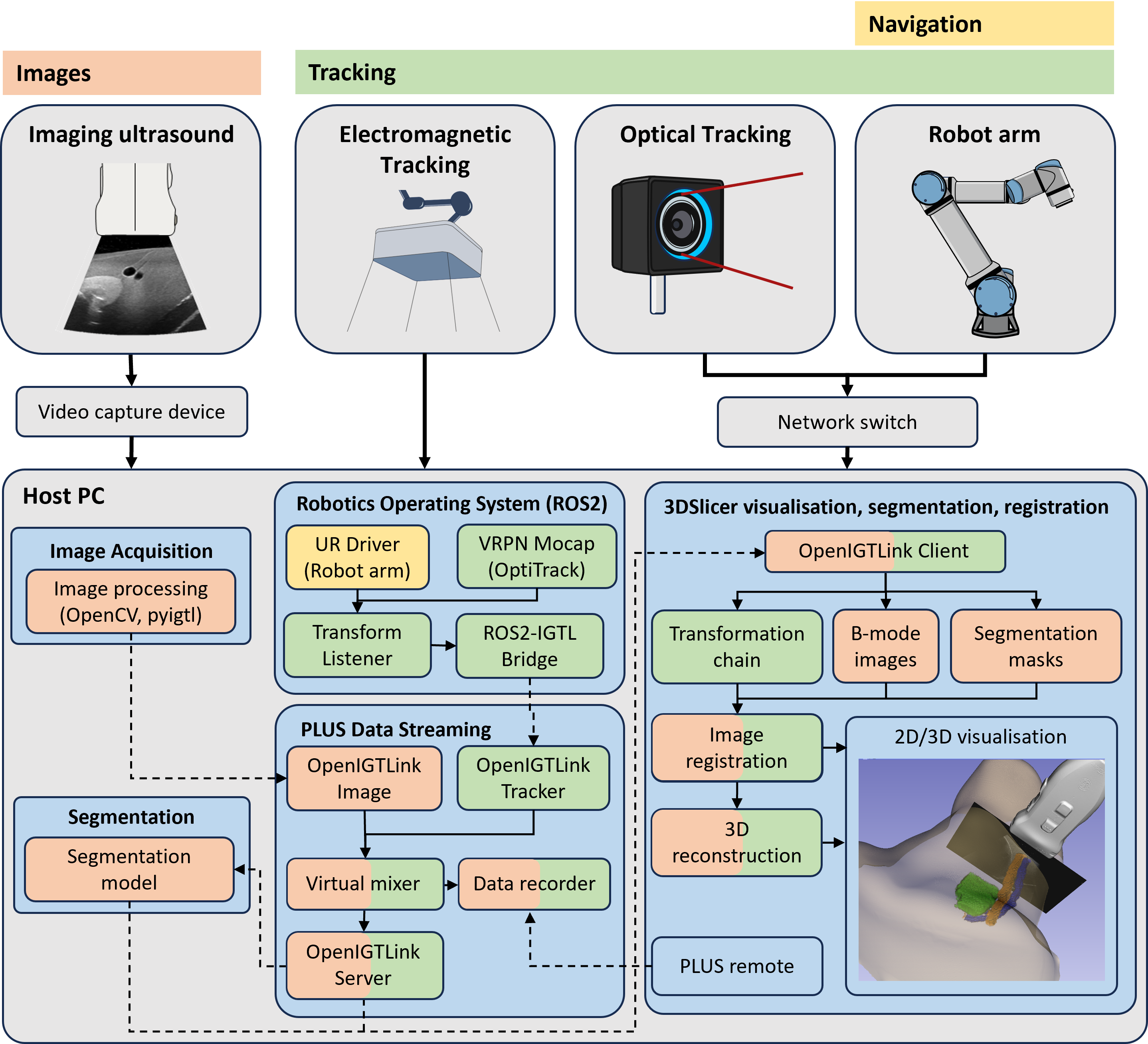}
\caption{Technical diagram showing hardware devices (grey) and software modules (blue) included as part of the experimental system used for real-time \ac{3D} \ac{US} reconstruction. The OpenIGTLink protocol was used to exchange data between modules (dashed arrows). The data flow within modules (thin solid arrows) and connections between hardware devices are also shown (thick solid arrows).}
\label{fig:technical_diagram}
\end{figure*}

The system uses open source tools to support integration with multiple imaging and tracking systems. Modules for data acquisition, robotic navigation, data streaming, segmentation, and visualisation ran on a distributed system using OpenIGTLink for real-time communication~\cite{tokuda_openigtlink_2009} (Fig.~\ref{fig:technical_diagram}). ROS2 (Foxy) interfaced with the robot and optical tracking systems, PLUS (v2.9.0) handled coordination synchronisation~\cite{lasso_plus_2014}, and 3DSlicer (v5.2.2) with SlicerIGT enabled visualisation, reconstruction, and analysis~\cite{ungi_open-source_2016}. Segmentation and \ac{3D} reconstruction were implemented in Python 3.9 on an Ubuntu 20.04 workstation (Intel i7, 32~GB RAM). Full instructions for the system setup, as well as relevant scripts, configuration files, and example code is included in the GitHub repository (https://github.com/SMRTUS/US-QA-3D).

\ac{US} images were acquired using a point-of-care \ac{US} system (GE, Venue Fit) and a L4‑20t‑RS transducer (3–20 MHz, 38.4 mm \ac{FOV}) and frames were captured with an HDMI-to-USB 3.0 capture card (Magewell, USB Capture HDMI Gen 2) and processed at the scanner's native resolution (1920x1080) and frame rate (15-30 FPS) using OpenCV. This acquisition approach is broadly compatible with modern \ac{US} systems providing HDMI output, though device-specific APIs may reduce latency. 

High-resolution harmonic B-mode imaging used 10/20 MHz frequency, 78 dB dynamic range, speckle reduction, and virtual convex mode to extend the \ac{FOV}. Using these imaging parameters, the axial, lateral, and elevational resolution (slice thickness) were 0.25-0.5~mm, 0.5-1.0~mm, and 0.7-2.0~mm respectively at 3~cm focal depth, measured using a multipurpose \ac{QA} phantom (CIRS, 040GSE)~\cite{computerized_imaging_reference_systems_inc_multi-purpose_2013}. 

A 6~\ac{DoF} collaborative robotic arm (Universal Robots, UR3e) was used to move and track the transducer pose. The transducer was secured in a custom two-part \ac{3D}-printed holder mounted onto the robot flange. The robot was interfaced via ROS2 and the Universal Robots drivers~\cite{universal_robots_universal_2025-1}, which provided real-time pose estimates based on the robot's internal kinematics. A custom ROS2 transform listener converted axis-angle-representation poses to transformation matrices and streamed them to PLUS via the ROS2-IGTL-bridge~\cite{connolly_bridging_2022}. The module MoveIt for ROS2 was used for motion planning~\cite{noauthor_moveitmoveit2_2025} to generate consistent scanning trajectories. Baseline \ac{QA} scans used linear translation at 5.0~mm/s, unless otherwise stated. 

The transducer's position and orientation were also captured using optical and \ac{EM} tracking. The optical tracking system used four infrared cameras (OptiTrack, Prime 13), positioned to maximise coverage of the working area, and calibrated using a standard wand-based dynamic calibration (OptiTrack Motive v2.2.0)~\cite{optitrack_naturalpoint_inc_calibration_2022}. A rigid-body frame with four 6.4~mm retroreflective markers was mounted to the transducer holder for passive tracking. A ROS2 VRPN client~\cite{sun_alvinsunyixiaovrpn_mocap_2025} broadcast timestamped poses at 60~FPS, and a custom ROS2 listener forwarded the timestamped transformations to PLUS via ROS2-IGTL-bridge~\cite{tobias_frank_openigtlinkros2_igtl_bridge_2024}. 

\ac{EM} tracking used a Planar 20-20 Field Generator, control unit, sensor interface unit, and 6-DoF sensor (NDI, Aurora), with manufacturer-reported positioning accuracy of 0.48~mm and orientation accuracy of 0.3$^{\circ}$~\cite{northern_digital_inc_minimally_2024}. The field generator covered a 500 mm$^3$ workspace and was positioned to minimise \ac{EMI}. The control unit was connected via USB 3.1, and timestamped poses were acquired directly through PLUS's native Aurora interface (Fig. \ref{fig:technical_diagram})~\cite{perk_lab_pluslib_2025}.

\subsection{Calibration and registration}
\label{sec:methods:calibration}
Static transforms defining the system geometry included the flange-to-transducer offset from \ac{CAD} measurements, robot linkage transforms from \ac{DH} parameters~\cite{universal_robots_dh_2026}, image-to-transducer transform from spatial calibration, and coordinate registration transforms. These were organised in 3DSlicer using the \ac{MRML} transform hierarchy, enabling consistent propagation of dynamic motion and tracing of \ac{US} image coordinates relative to the robot base. 

Temporal and spatial calibration were performed using the PLUS freehand calibration tool (fCal)~\cite{lasso_plus_2014}. Temporal calibration estimated the latency between the image and tracking streams by imaging the bottom of a water tank while moving the transducer in a periodic oscillatory motion. Spatial calibration determined the static transform between the \ac{US} image and the transducer (ImageToTransducer) using a 3-layer N-wire phantom~\cite{carbajal_improving_2013}, giving a sub-millimetre registration error. 

All tracking modalities (robotic, optical, and \ac{EM}) were aligned to a common world coordinate system defined by the robot base frame. Rigid transforms between coordinate-tracking systems were computed via fiducial registration using at least 8 reference points across the workspace. The 3DSlicer volume reslice driver rendered the \ac{US} image within the \ac{3D} scene during scanning.

\subsection{Segmentation and 3D reconstruction}
\label{sec:methods:segmentation}
Real-time \ac{3D} reconstruction combined frame-by-frame segmentation with voxel-based volume mapping, providing live visual feedback. Segmentation ran in parallel with the imaging pipeline, so the latency from model inference did not bottleneck visualisation. Tracked frames were processed to produce segmentation masks which retained the original pose and timestamp for synchronised \ac{3D} reconstruction. 

A threshold-based model segmented anechoic targets in B-mode images of the phantom for \ac{QA} (D). Processing steps included \ac{ROI} cropping, median filtering (kernel size 7) to reduce speckle noise while preserving edges, and Otsu's thresholding to create binary masks~\cite{otsu_threshold_1979, russ_image_2018}. Masks were refined using morphological closing to fill gaps and morphological opening to remove small islands. A trimming function removed dark vertical regions from poor coupling. The resulting binary masks were then used as the input for \ac{3D} reconstruction. 

Real-time \ac{3D} reconstruction used the PLUS Volume Reconstruction module in 3DSlicer (SlicerIGT)~\cite{ungi_open-source_2016}, with voxel-based forward mapping to transform each \ac{2D} pixel into a \ac{3D} voxel coordinate, and nearest neighbour interpolation to assign voxel values within a pre-defined grid~\cite{mohamed_survey_2019}. When multiple pixels mapped to the same voxel, the final value was determined by maximum-value compounding. 

Volumes were updated continuously and rendered using 3DSlicer's \ac{VTK} GPU Ray Casting, enabling real-time \ac{2D} and \ac{3D} visualisation~\cite{bozorgi_gpu-based_2015}. Segmentation overlays were displayed on synchronised B-mode images, and multiplanar formatting allowed for axial, sagittal, coronal, or oblique views of the resulting volumes. This allowed for clear visualisation of anatomical details, aligned with patient anatomy or procedural targets, regardless of the original scan orientation~\cite{mohamed_survey_2019}. 

\subsection{Evaluation and Quality Assurance}
\label{sec:methods:QA}

To evaluate reconstruction accuracy and repeatability, a purpose-built phantom for \ac{US-QA-3D} was developed. The phantom contained four inclusions with known geometry: a sphere, an ellipsoid, a cylinder, and an equilateral triangular prism, selected for their varied geometric properties and relevance to anatomical structures. The inclusions were cast from an agar mixture without acoustic scatterers, producing anechoic targets whose dimensions serve as ground truth for \ac{3D} reconstruction. Inclusions were embedded in a tissue-mimicking agar block containing polydisperse glass beads to introduce speckle noise on B-mode, creating clear contrast against the anechoic targets. Phantom preparation details are provided in Appendix \ref{appendix:phantom}.

\begin{figure*}[htbp]
\centering
\includegraphics[width=\textwidth]{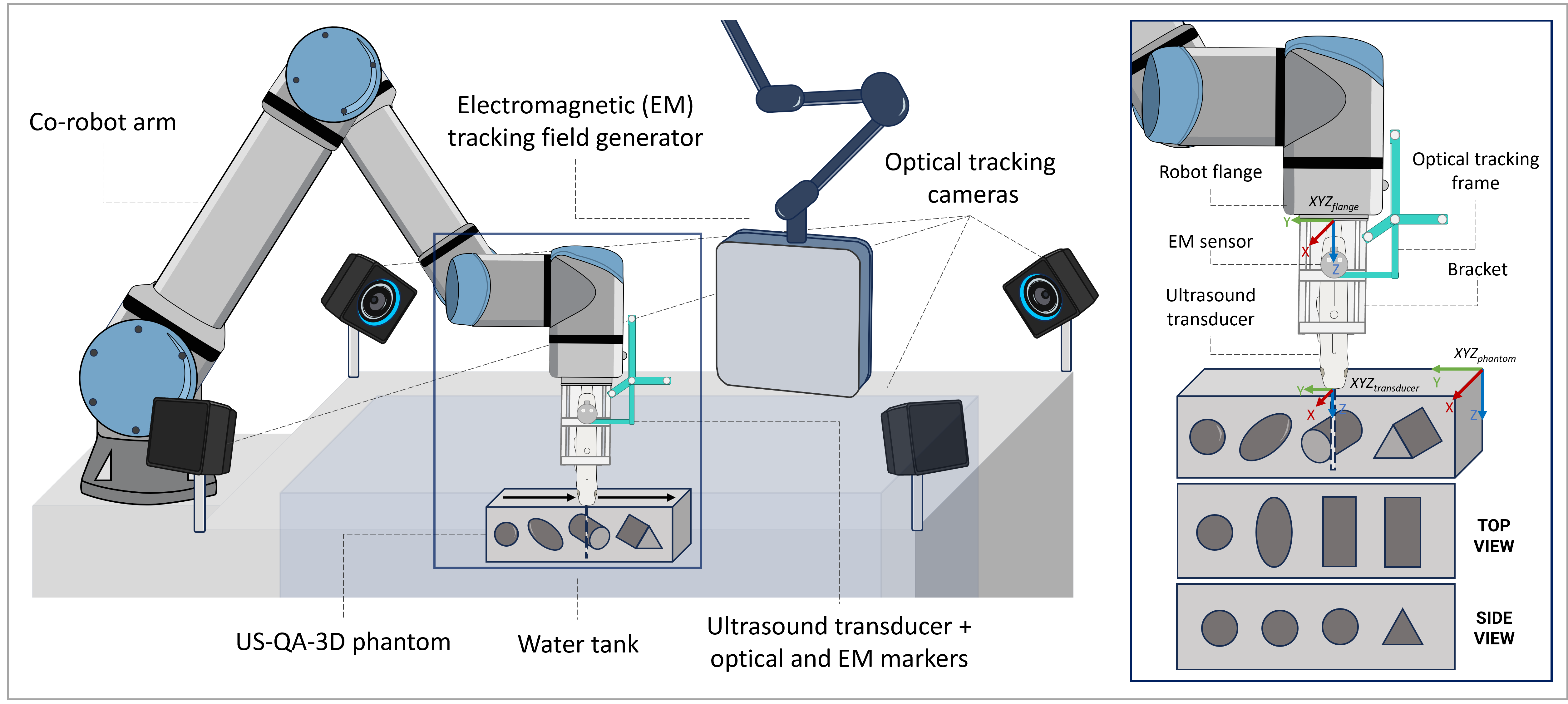}
\label{fig:experimental_setup}
\caption{Experimental setup for experiments with \ac{US-QA-3D} phantom, showing the robotic arm, ultrasound transducer, tracking assembly, water tank with geometric phantom, optical cameras, and electromagnetic field generator. The co-robotic arm was used for navigation. An EM sensor and an optical tracking frame enable simultaneous multimodal tracking alongside the robot's kinematics. The reference frames for the robot flange ($\textit{XYZ}_\textit{flange}$), transducer ($\textit{XYZ}_\textit{trasnducer}$), and phantom ($\textit{XYZ}_\textit{phantom}$) are shown. The \ac{US-QA-3D} phantom with geometric inclusions is shown in oblique, top, and side views.}
\end{figure*}

\ac{QA} experiments were conducted with the phantom suspended in a large water tank filled with filtered, degassed water. This ensured consistent acoustic coupling and unrestricted robot motion. The phantom was scanned using robotic navigation with simultaneous \ac{EM}, optical, and robotic kinematic tracking (Fig.~\ref{fig:experimental_setup}). Optical cameras were arranged to maximise marker visibility, and the \ac{EM} field generator was positioned to minimise \ac{EMI}, without restricting probe motion. A $\sim$1~cm water layer separated the transducer from the phantom to prevent target deformation. 

Baseline scans used automated linear trajectories at 5.0~mm/s, with the imaging plane orthogonal to the phantom's long axis. Additional experiments varied scan speed (2.5–17.5~mm/s) and insonation angles (30°, 45°, and 90° about the axial axis and $\pm$~20° about the lateral axis). At large axial angles, multi-sweep scans were required due to the limited B-mode \ac{FOV}, and overlapping regions were combined using maximum-voxel compounding. Image and tracking data were recorded through PLUS and triggered via the PLUS remote for 3DSlicer.

Reconstructed volumes were analysed using a Python pipeline with the 3DSlicer \ac{API}. Intensity thresholding and connected-component analysis separated the four geometric targets, after which geometric descriptors were calculated, including the volume, surface area, centroid location, Feret diameter, roundness, flatness, elongation, and principal axes of variation~\cite{russ_image_2018}. 

Reference models derived from physical measurements of the agar objects were reconstructed in 3DSlicer \ac{VTK} and co-registered with the corresponding segmented shapes using \ac{ICP} registration. For quantitative evaluation, 3DSlicer's Segment Comparison and Segment Statistics modules were used. Agreement between predicted ($X$) and reference ($Y$) geometries was measured using the \ac{DSC}:
\begin{equation}
 \textrm{DSC} =\frac{2\left | X\cap Y\right |}{\left | X\right |+\left | Y\right |}
\end{equation}
In \ac{2D}, \ac{DSC} measured the pixel-wise agreement between predicted and manually annotated segmentation masks, whereas in \ac{3D} (DSC-3D), it quantified voxel-wise agreement between reconstructed and reference volumes. 

Boundary accuracy was assessed using the \ac{HD}, which measures the difference between the surfaces of the predicted shape ($X_s$) and reference shape ($Y_s$). For each surface point ($x\in X_s$) and ($y\in Y_s$), the set of shortest distances to the corresponding shape ($d_{X_s\rightarrow Y_s}$ and $d_{Y_s\rightarrow X_s}$) was calculated: 
\begin{equation}
  d_{X_s\rightarrow Y_s} =\left\{\min_{y\in Y_s}\| x - y\|\; :\; x\in X_s\right\}
\end{equation}
\begin{equation}
  d_{Y_s\rightarrow X_s} =\left\{\min_{x\in X_s}\| y - x\|\; :\; y\in Y_s\right\}
\end{equation}
This was used to find the (bidirectional) maximum Hausdorff Distance ($\text{HD}_{\text{max}}$), defined by the maximum of both sets of shortest distances:  
\begin{equation}
 \text{HD}(X_s, Y_s) =\max\left(\max (d_{X_s\rightarrow Y_s}),\;\max (d_{Y_s\rightarrow X_s})\right)
\end{equation}
and the 95th-percentile \ac{HD} ($\text{HD}_{95}$):
\begin{equation}
 \text{HD}_{95}(X, Y) =\max\left( P_{95}(d_{X\rightarrow Y}),\; P_{95}(d_{Y\rightarrow X})\right)
\end{equation}
Both the \ac{HD} and \ac{HD95} are positive values expressed in millimetres, with 0 mm indicating a perfect prediction. 

To complement voxel-based metrics, shape-fitting provided geometry-specific information. Surface meshes were converted to point clouds and fitted using SciPy optimisation to estimate the sphere centre and radius; ellipsoid centre and three semi-axes; cylinder centre, radius, height, and axis; and triangular prism height, axis orientation, and base vertices.

\section{Results}

\subsection{System Performance}

Our system for tracked \ac{US} reconstruction demonstrated reliable, real-time performance in multimodal tracking \ac{QA} experiments (Table~\ref{tbl:system_performance}). Tight synchronisation and low system latency were essential for accurate real-time \ac{3D} reconstruction and visualisation. Frame-by-frame synchronisation was handled using PLUS's internal buffering, with a local time offset uncertainty of $<$~0.5~ms, as measured by the FCal temporal calibration algorithm. End-to-end latency from \ac{US} capture to visualisation was approximately 90.2~$\pm$~7.4~ms, consistent with expected delays for image transfer and display using PLUS/OpenIGTLink and 3DSlicer~\cite{lasso_plus_2014}. Volume reconstruction and \ac{VTK} rendering allowed live updates every 0.2s.

Accurate tracking and registration were also critical to achieve sub-millimetre precision in the output volumes. Image-to-transducer calibration with the N-wire technique gave a \ac{RMS} error of 0.81~$\pm$~0.15~mm, consistent with the reported performance of this technique~\cite{carbajal_improving_2013}. Robotic kinematic tracking using the robot arm's internal joint transformations yielded highly accurate end-effector tracking, with reported pose positioning errors of $<$~0.03~mm. Optical tracking calibration achieved a mean reprojection \ac{3D} error of $<$~0.1~mm across all experiments, but tracking was affected by marker visibility, leading to gaps in the reconstructed volumes when the line of sight was obstructed. \ac{EM} tracking accuracy was limited by field distortion, despite careful placement of the \ac{EM} field generator to minimise it. The \ac{RMS} error was 0.1-0.2~mm in controlled conditions but increased depending on the tracker's position relative to the robot, sometimes exceeding several millimetres. Spatial alignment across tracking modalities had a mean \ac{RMS} registration error of 0.19~$\pm$~0.07~mm. 

\begin{table}[htbp]
\footnotesize
\setlength{\tabcolsep}{3pt}
\centering
\caption{System performance, including timing and synchronisation, calibration and registration, and tracking accuracy.} 
\label{tbl:system_performance}
\begin{tabular}{cll}
\toprule
\multicolumn{1}{l}{}                                                                     & Metric                                & Value             \\ \hline
\multirow{3}{*}{\begin{tabular}[c]{@{}c@{}}System \\ Timing\end{tabular}}                & Synchronisation uncertainty           & \textless~0.5 ms  \\
                                                                                         & End-to-end latency                    & 90.2 ± 7.4 ms     \\
                                                                                         & Volume update rate                    & 0.2 s             \\ \hline
\multirow{2}{*}{\begin{tabular}[c]{@{}c@{}}Calibration and \\ registration\end{tabular}} & Image-to-transducer RMS error         & 0.81 ± 0.15 mm    \\
                                                                                         & Cross-modality RMS registration error & 0.19 ± 0.07 mm    \\ \hline
\multirow{3}{*}{\begin{tabular}[c]{@{}c@{}}Tracking\\ Accuracy\end{tabular}}             & Robot kinematic tracking accuracy     & \textless~0.03 mm \\
                                                                                         & Optical tracking reprojection error   & \textless~0.1 mm  \\
                                                                                         & EM tracking RMS error                 & 0.1 – 0.2 mm      \\ \hline
\end{tabular}
\end{table}
\vspace{-20pt}

\subsection{Geometric phantom}

Evaluation using the \ac{US-QA-3D} phantom enabled controlled assessment of reconstruction accuracy, confirming the system's ability to reconstruct known \ac{3D} shapes with low error. Combined, voxel-, surface-, shape-, and fit-based metrics provided a comprehensive evaluation of system performance under controlled conditions and highlighted the importance of accurate tracking and consistent scanning for high-quality \ac{3D} reconstruction.

\begin{figure}[htbp]
\centering
\includegraphics[width=\textwidth]{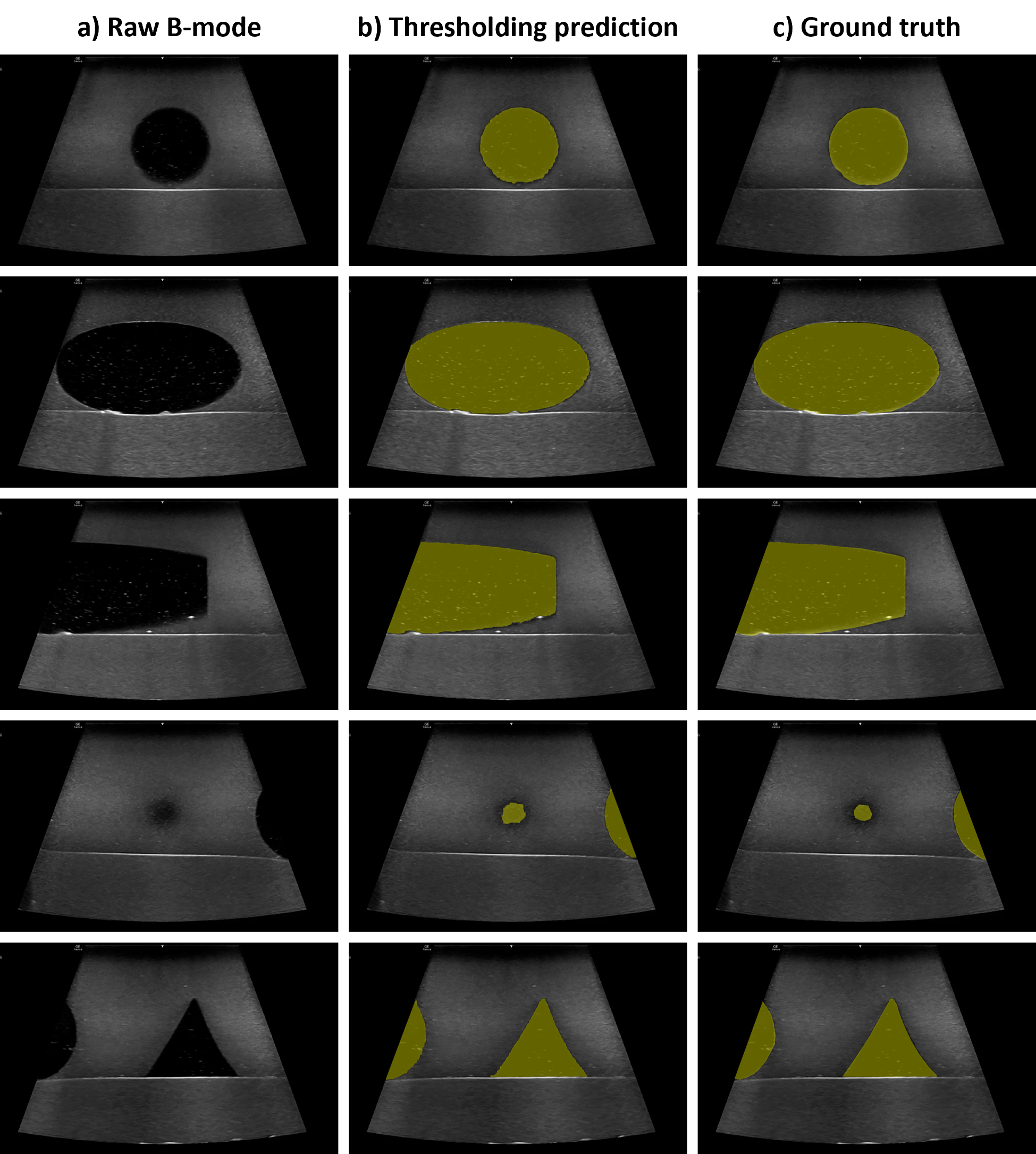}
\caption{Thresholding-based segmentation of objects in the \ac{US-QA-3D} phantom. Comparison shows a) B-mode images, b) B-mode images with overlaid prediction mask, and c) B-mode images with overlaid ground-truth reference mask. 
}
\label{fig:seg_geom}
\end{figure}

Construction of the \ac{US-QA-3D} phantom was straightforward, and B-mode imaging showed good contrast between the anechoic inclusions and speckle pattern of the outer media (Fig.~\ref{fig:seg_geom}a). The segmentation algorithm was tuned to define the \ac{ROI}, segment the inclusions, and optimise kernels for post-processing. Small bubbles introduced to the agar inclusions during manufacturing did not affect segmentation, as closed-contour filling reliably removed them. Automatic thresholding-based segmentation delineated the inclusions accurately (Fig.~\ref{fig:seg_geom}b). Comparison with hand-annotated ground truth images (Fig.~\ref{fig:seg_geom}c) showed excellent agreement, with a mean \ac{DSC} of 0.97 on a test dataset of 50 images. The thresholding approach was also computationally efficient, with a processing time of 21.8~$\pm$~6.4~ms on a mid-range processor. 

\begin{figure}[htbp]
\centering
\includegraphics[width=\textwidth]{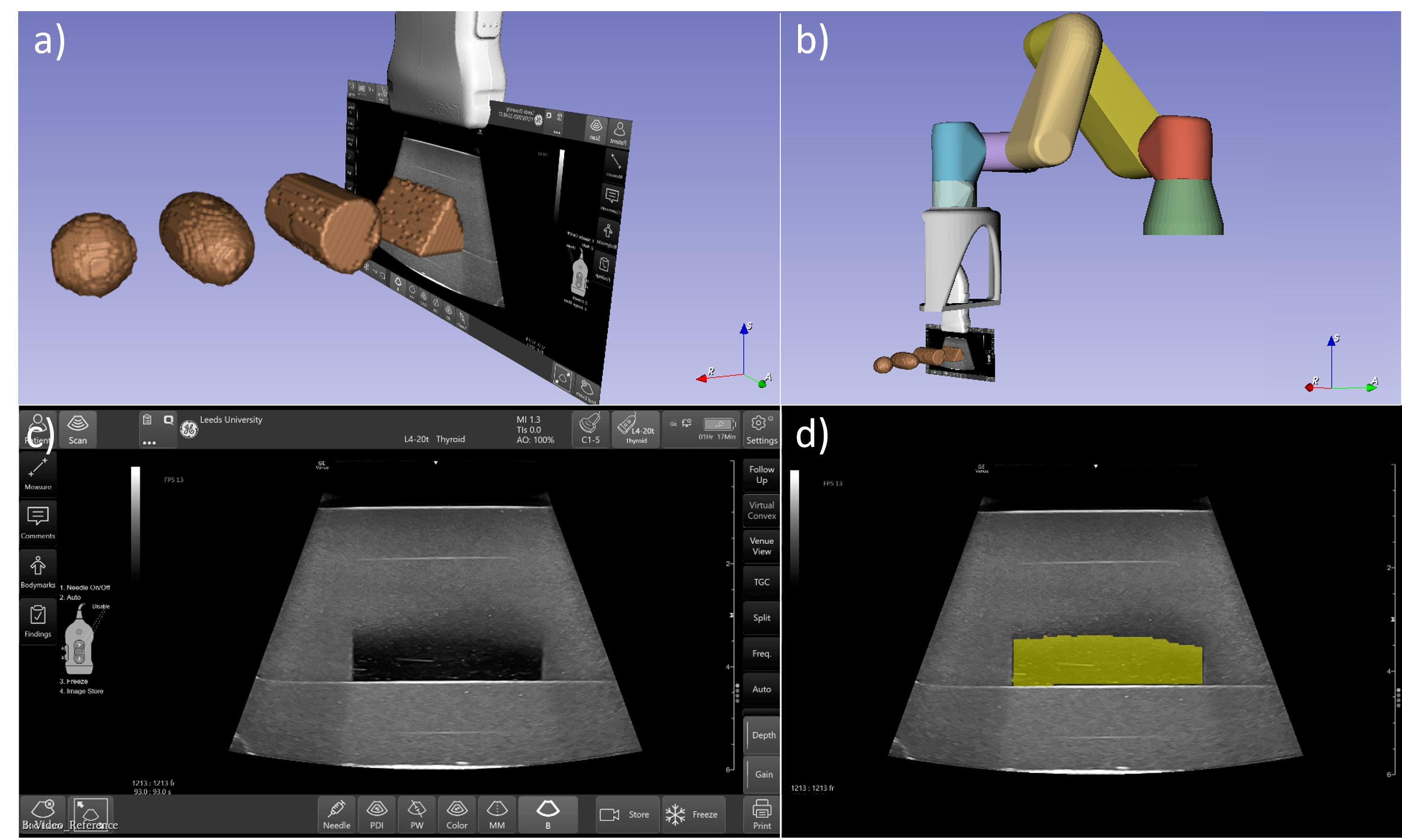}
\caption{Dual \ac{3D} view of live real-time volumetric reconstruction of robotic-tracked \ac{US} in the \ac{US-QA-3D} phantom. a) Close-up of voxel-based volume reconstruction showing the tracked \ac{US} frame and reconstructed geometry. b) Robot arm and transducer assembly during the scan. c) B-mode image from the \ac{US} machine showing a cross-section of the triangular prism inclusion, which appears anechoic compared to the background. d) B-mode image with segmentation overlay from the thresholding algorithm. For a video of the live \ac{3D} reconstruction, see Supplementary Video.}  
\label{fig:geom_reconstruction}

\end{figure}

The segmentation model was applied for real-time \ac{3D}-tracked \ac{US} reconstruction with robotic navigation and simultaneous optical, \ac{EM}, and kinematic tracking. During scanning, reconstructed volumes were rendered and updated in real time (Supplementary Video), allowing the operator to confirm complete coverage of the \ac{ROI} (Fig.~\ref{fig:geom_reconstruction}a). A second \ac{3D} view displayed the robot, transducer, and tracked \ac{US} frames (Fig.~\ref{fig:geom_reconstruction}b). The live \ac{US} machine feed and B-mode images were shown in a dedicated \ac{2D} view (Fig.~\ref{fig:geom_reconstruction}c), while a second \ac{2D} view displayed the segmentation overlay (Fig.~\ref{fig:geom_reconstruction}d). 

\begin{figure*}[htbp]
\centering
\includegraphics[width=\textwidth]{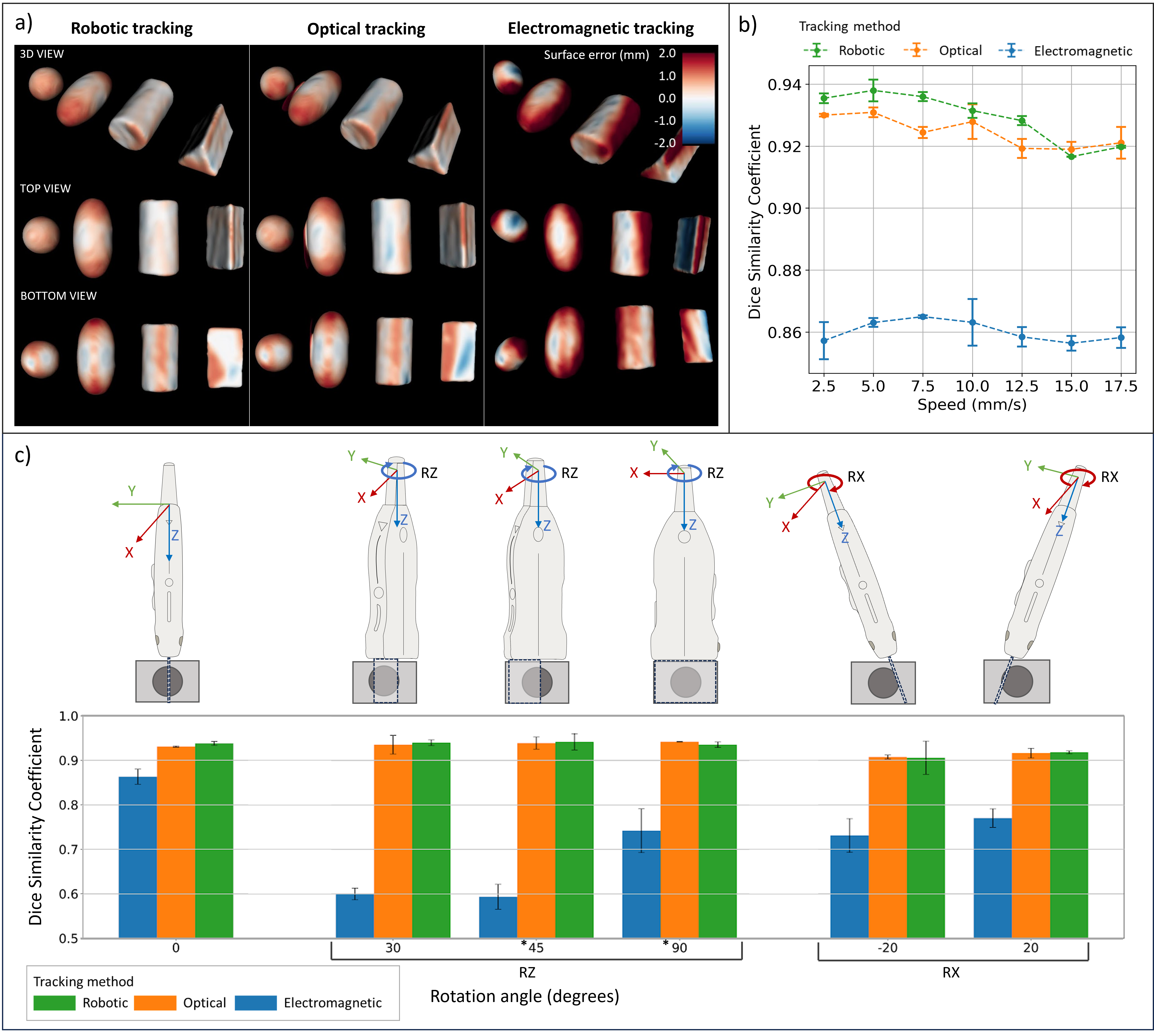}
\caption{\ac{3D} reconstruction of objects in \ac{US-QA-3D} phantom for different tracking modalities (robotic, optical, and electromagnetic). a) Heat maps show the surface deviation error compared with co-registered reference objects. b) Mean DSC across shapes at different scanning speeds. c) Mean DSC at varying insonation angles. Error bars show the standard deviation across three repeats. *These experiments required multi-sweep (compound) scanning.
}
\label{fig:us-qa-3d-grouped}
\end{figure*}

Baseline reconstruction quality was evaluated using segmentation-, shape-, and fitting-based metrics, averaged across the four geometric targets (Table~\ref{tbl:us-qa-3d}). Robotic kinematic and optical tracking performed well, with mean DSC-3D of 0.94~$\pm$~0.01 and 0.93~$\pm$~0.01, respectively, while \ac{EM}-based reconstruction was poorer, with DSC-3D of 0.86~$\pm$~0.02, likely due to the influence of \ac{EMI}. This is also supported by the \ac{HD} and \ac{HD95}, which measure the surface fidelity of the reconstructed shapes. For robotic kinematic and optical tracking, the maximum \ac{HD} was $<$~2~mm and \ac{HD95} remained $<$~1.2~mm, but \ac{EM}-tracked reconstruction had a \ac{HD95} $>$~2~mm. Fig.~\ref{fig:us-qa-3d-grouped}a shows segmented surfaces from the same imaging sequence using each tracking method, coloured by surface error relative to the co-registered ground truth.

\begin{table*}[ht]
\footnotesize
\setlength{\tabcolsep}{3pt}
\centering
\caption{Baseline scans of the \ac{US-QA-3D} phantom, using simultaneous tracking with robotic kinematic, optical and electromagnetic (EM) tracking. Fitting metrics provide the following shape measurements: sphere radius (R); ellipsoid semi-major axis (R1) and semi-minor axis (R2) radii; cylinder radius (R) and height (H); and triangular prism length (L) and height (H). All metrics are given as mean $\pm$ standard deviation. The error relative to the ground truth is shown in brackets for the fitting metrics.} 
\label{tbl:us-qa-3d}
\begin{tabular*}{\textwidth}{@{\extracolsep{\fill}} c c l c c c c @{}}
\toprule
\multicolumn{2}{l}{}                                            & Robotic              & Optical              & EM                  \\\midrule
\multirow{3}{*}{Segmentation metrics} & DSC-3D                     & 0.94 ± 0.01          & 0.93 ± 0.01          & 0.86 ± 0.02         \\
                                      & HD (mm)                 & 1.79 ± 0.37          & 1.93 ± 0.73          & 3.87 ± 0.87         \\
                                      & HD95 (mm)               & 1.17 ± 0.12          & 1.19 ± 0.12          & 2.43 ± 0.44         \\\midrule
\multirow{4}{*}{Shape metrics}        & Volume error (\%)       & 11.80 ± 1.73         & 11.90 ± 1.47         & 20.05 ± 4.77        \\
                                      & Surface Area error (\%) & 8.90 ± 2.33          & 10.41 ± 6.76         & 18.10 ± 8.54        \\
                                      & Sphere roundness        & 0.98 ± 0.01          & 0.98 ± 0.01          & 0.95 ± 0.01         \\
                                      & Ellipsoid elongation    & 2.08 ± 0.01          & 2.07 ± 0.04          & 1.80 ± 0.02         \\\midrule
\multirow{7}{*}{Fitting metrics}      & Sphere R (mm)           & 12.05 ± 0.03 (+0.48) & 12.07 ± 0.01 (+0.50) & 12.13 ± 0.09 (+0.56)~\\
                                      & Ellipsoid R1 (mm)       & 12.73 ± 0.09 (+0.40) & 12.83 ± 0.37 (+0.50) & 14.75 ± 0.11 (+2.42)\\
                                      & Ellipsoid R2 (mm)       & 26.56 ± 0.11 (+1.91) & 26.59 ± 0.07 (+1.94) & 26.74 ± 0.13 (+2.09)\\
                                      & Cylinder R (mm)         & 11.23 ± 0.08 (-0.73) & 11.14 ± 0.08 (-0.82) & 11.93 ± 0.26 (-0.03)\\
                                      & Cylinder H (mm)         & 44.94 ± 0.40 (+2.37) & 45.14 ± 0.13 (+2.57) & 45.00 ± 0.46 (+2.43)\\
                                      & Triangular prism L (mm) & 22.33 ± 0.47 (-1.18) & 22.42 ± 0.26 (-1.09) & 24.87 ± 0.96 (+1.36)\\
                                      & Triangular prism H (mm) & 38.54 ± 0.07 (+1.56) & 38.71 ± 0.35 (+1.73) & 40.69 ± 0.85 (+3.71)\\\bottomrule
\end{tabular*}
\end{table*}

Shape-based metrics provided additional validation on the geometric reconstruction accuracy. The roundness of the segmented sphere was 0.98~$\pm$~0.01 for both robotic and optical tracking, close to the expected value of 1.0. The ellipsoid elongation was 2.07~$\pm$~0.01 for robotic tracking and 2.06~$\pm$~0.01 for optical tracking, both close to the expected value of 2.0. \ac{EM}-tracked reconstructions were visually poorer, with a sphere roundness of 0.95~$\pm$~0.01 and ellipsoid elongation of 1.80~$\pm$~0.02 indicating shape distortion. 

Fit-based shape analysis also demonstrated low parameter errors for sphere and ellipsoid models under robotic and optical tracking. Fitted radii deviated by less than 0.5~mm for the sphere Radius $R$, and for the ellipsoid semi-axis radii $R_y$ and $R_z$, but were slightly higher for $R_x$, potentially due to variation in lateral image resolution across the wide virtual convex imaging window. The cylinder and triangular prism also exhibited slightly higher parameter errors for dimensions along the lateral axis (x-axis), with fitted dimensions up to 2.57~mm larger than the measured values (for the cylinder reconstructed with optical tracking). Despite this, all shape-fitting functions converged successfully with low residual error. 

The largest errors were observed when segmentation quality was poor, particularly in cases of surface dropout, which was occasionally seen with \ac{EM} tracking. Depending on the scan orientation, shapes were also slightly elongated in the elevational direction, resulting in errors up to 1~mm in the fitted shape parameters. This was likely due to the transducer's elevation thickness (0.7-2.0~mm), which causes signal spill-out from and out-of-plane elongation due to contributions above or below the imaging plane. In most cases, the shape-fitting parameters slightly exceeded the reference values, indicating a small bias toward overestimating the size of the reference objects.

Comparison with resolution-limited theoretical maximums indicated that reconstruction accuracy was largely constrained by \ac{US} imaging rather than calibration, tracking, synchronisation, or segmentation. For example, given a reference sphere of radius $r$, the volume $V$ is
\begin{equation}
    V = 4/3\cdot\pi r^3,
\end{equation}
and a reconstructed sphere with radius error $r'$, the volume $V'$ is
\begin{equation}
    V' = 4/3\cdot\pi(r+r')^3.
\end{equation}
If the spheres are perfectly aligned and concentric, the theoretical resolution-limited DSC-3D is
\begin{equation}
    \frac{2V}{V+V'} = \frac{2r^3}{r^3+(r+r')^3}
\end{equation}
In the \ac{US-QA-3D} phantom, the sphere had radius $r=11.5$~mm and a resolution-limited uncertainty $r'=0.5$~mm, which gives an expected maximum DSC-3D of 0.94, closely matching the experimental values of 0.94~$\pm$~0.01 for robotic, and 0.93~$\pm$~0.01 for optical-tracked \ac{3D} reconstruction. Similarly, the theoretical resolution-limited sphere volume error is approximately 12\%, close to the experimental values of 11.80~$\pm$~1.73 and 11.90~$\pm$~1.47.

Best results were obtained using slow, constant scanning speeds, with slight degradation in reconstruction quality when the scanning speed exceeded 7.5~mm/s (Fig.~\ref{fig:us-qa-3d-grouped}b). At speeds $\leq$ 7.5~mm/s, robotic tracking performed as well as or slightly better than optical tracking and was much better than \ac{EM} tracking. Under these conditions, reconstruction was both accurate and repeatable, with a mean DSC-3D $>$ 0.92 and \ac{HD95}~$<$~1.2~mm. At higher navigation speeds, reconstruction quality remained acceptable but decreased slightly for both robotic and optical tracking, likely due to minor synchronisation errors producing surface unevenness and small gaps. Overall, robotic tracking produced the most consistent reconstructions, followed closely by optical tracking, while \ac{EM} tracking performed worse when used alongside the robot arm.

For robotic and optical tracking, insonation angle had little or no effect on the mean DSC-3D, even during multi-sweep scanning (Fig.~\ref{fig:us-qa-3d-grouped}c). Axial rotations (z-axis) produced accurate reconstructions despite the non-zero angle between the normal vector of the imaging plane and the scanning direction. For rotations $>$ 30°, maximum-value compounding enabled complete reconstruction, with no clear demarcation between the sections from multi-sweep scanning. Rotations about the lateral axis (x-axis) introduced an oblique angle between the transducer face and the phantom surface, increasing artefacts from reflection and refraction and slightly reducing the DSC-3D compared with perpendicular scanning. Across all experiments, the standard deviation between repeats was small, demonstrating good repeatability. In \ac{EM} scanning experiments, the DSC-3D of the reconstructed shape was more angle-dependent since field distortion effects varied with the relative angle between the sensor and the scanner, as well as the robot position.

\section{Discussion and Conclusion}

This study presents a flexible, open source tracked \ac{US} platform for \ac{3D} reconstruction and a standardised \ac{QA} framework for quantitative system validation. We demonstrate that accurate volumetric reconstruction can be achieved with a conventional \ac{2D} B-mode transducer combined with interchangeable pose tracking using robotic kinematic, optical, or \ac{EM} tracking. The platform also supports simultaneous multimodal tracking, which enables co‑registered applications with other tracked tools or external systems.  

Compared with existing phantom-based validation approaches designed for dedicated \ac{3D} transducers, the \ac{US-QA-3D} framework evaluates volumetric shape reconstruction to expose clinically relevant failure modes of tracked \ac{US}, including shape distortion, boundary noise, and scaling errors~\cite{iommi_evaluation_2020}. The \ac{US-QA-3D} phantom incorporates shapes with different symmetry properties, helping to identify errors from calibration, temporal misalignment of data streams, anisotropic distortion in tracking, and reconstruction artefacts that may not be visible in point-based tests. The phantom was inexpensive and simple to construct, and the geometry of the inclusions could be adapted to different applications. Using threshold-based segmentation and automatic registration enabled simple \ac{QA} without GPUs or specialised hardware, further simplifying implementation. Segmentation of B-mode images achieved high accuracy (\ac{DSC} = 0.97) and real-time performance (45~FPS), demonstrating its suitability for assessing errors associated with tracking and \ac{3D} reconstruction. This effectively decouples tracking and reconstruction errors from segmentation errors, a distinction often overlooked in prior studies. By separating these components, the \ac{QA} framework enables meaningful comparisons for different systems, transducers, and tracking modalities and provides a practical tool to accelerate future development.  

Applying this framework showed that both robotic kinematic tracking and optical tracking can achieve shape reconstruction close to the physics-based resolution limits of the \ac{US} transducer (DSC-3D = 0.94, \ac{HD95} = 1.2 mm). This is significant because optical tracking is highly accurate, with widely reported sub-millimetre \ac{RMS} positional accuracy~\cite{aurand_accuracy_2017}. Demonstrating performance comparable to optical tracking confirms that modern 6-DOF robots can provide sufficiently accurate pose tracking for high-quality \ac{3D} \ac{US}, while eliminating the line-of-sight issues and workflow complexity associated with optical cameras. The \ac{QA} also clarified the role of \ac{EM} tracking in tracked \ac{US}, which performed well in freehand calibration, but poorly when used alongside the robot arm. This is consistent with the well-documented susceptibility of \ac{EM} tracking to magnetic field distortion~\cite{poulin_interference_2002} and highlights the importance of volumetric \ac{QA} when using \ac{EM} tracking in clinical settings, where ferromagnetic objects and \ac{EMI} sources are often unavoidable~\cite{poulin_interference_2002}. 

Nevertheless, there are important limitations in this study that should form the basis for future research. This study used a single high-frequency linear probe, and further investigation is needed to evaluate performance with other transducers, systems, and imaging settings. Furthermore, while the \ac{QA} framework assessed geometric reconstruction under controlled conditions, future work should investigate robustness in non-idealised settings to quantify the additional effect of tissue deformation, acoustic shadowing, and patient motion on \ac{3D} reconstruction. Finally, although optical, \ac{EM}, and robotic kinematic tracking have been the dominant techniques for \ac{US} tracking, recent trackerless methods have also been developed that use \ac{AI} models to estimate transducer trajectories from video clips~\cite{peng_recent_2022, zhou_deep_2025, yandong_establishment_2024, wein_three-dimensional_2020}. These methods, while promising, were not assessed here as they remain difficult to validate quantitatively and are not yet suitable for safety-critical image-guided interventions. 

By standardising volumetric \ac{QA} for tracked \ac{US}, this work supports transparent performance reporting and facilitates cross-system comparison and evidence-based development of \ac{3D} \ac{US}. The proposed system showed accurate \ac{3D} reconstruction with robotic, optical, and \ac{EM} tracking which makes it extensible to a range of previously investigated clinical applications, including \ac{3D} reconstruction for organ volumetry~\cite{zielke_rsv_2022}, vascular assessment to determine the severity of stenosis~\cite{merouche_robotic_2016, chung_freehand_2017, almuhanna_carotid_2015}, and \ac{HIFU} treatment planning for tumour ablation ~\cite{daunizeau_robot-assisted_2020}. Ultimately, the framework will improve confidence in the reliability of \ac{3D} \ac{US} reconstruction, which is essential for safe and effective translation in clinical practice.

\section{Acknowledgments}
\noindent This study has received funding from the Engineering and Physical Sciences Research Council (EPSRC) through the following awards: EP/S024336/1 and EP/V04799X/1. J. Chandler was supported by the Leverhulme Trust and the Royal Academy of Engineering under a RAEng/Leverhulme Trust Research Fellowship (LTRF-2425-21-154).

\section{Appendix}

\subsection{US-QA-3D phantom fabrication}
\label{appendix:phantom}

\begin{figure}[!htbp]
\centering
\includegraphics[width=4.5in]{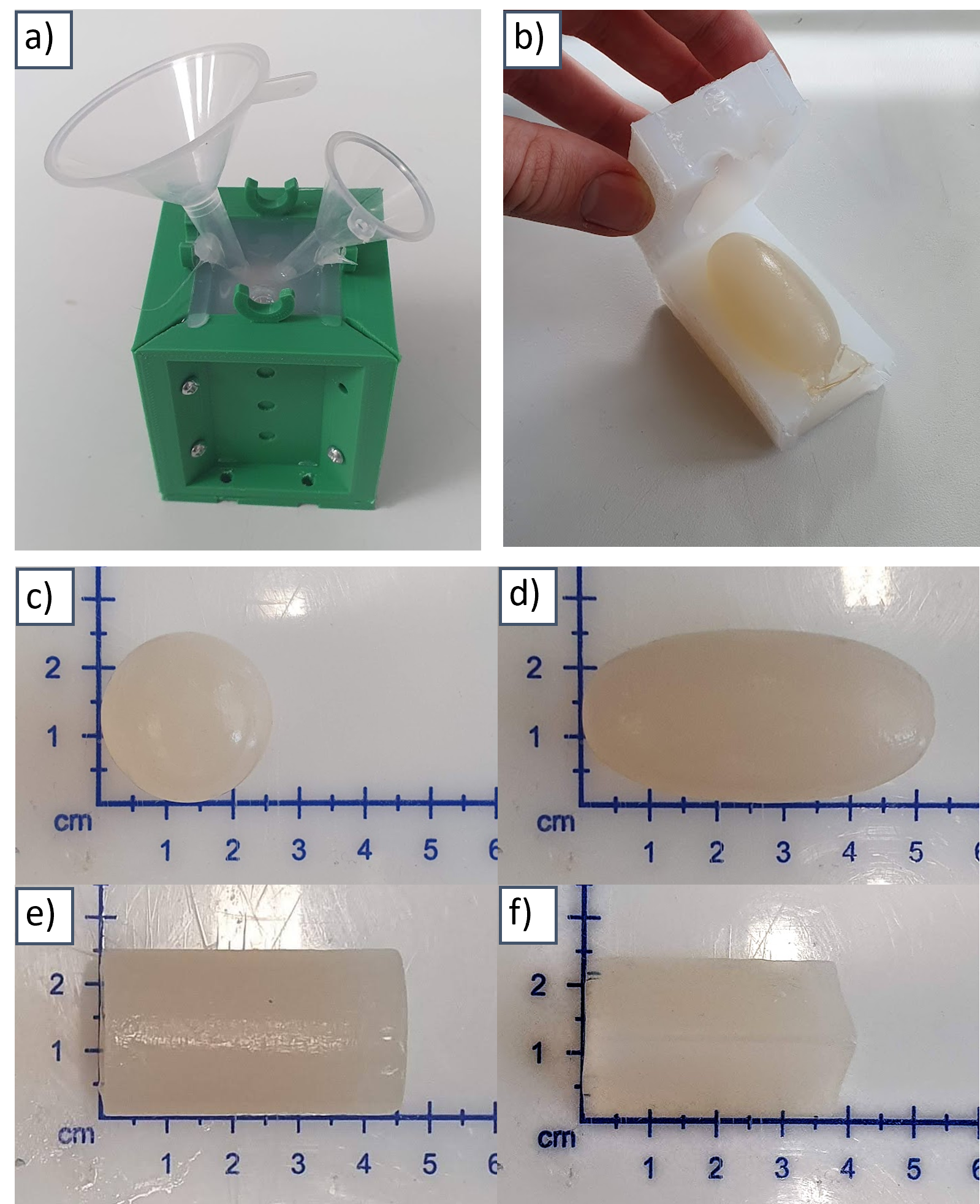}
\caption{\ac{US-QA-3D} phantom construction showing a) fabrication of silicone mould of reference object in \ac{3D}-printed container, b) silicone mould with agar inclusion (before trimming the sprue), c-f) agar sphere, ellipsoid, cylinder, and triangular prism reference objects.}
\label{fig:phantom}
\end{figure}

Silicone moulds of reference objects were created by casting negative impressions of wooden masters. Moulds were made of platinum-cure silicone (Smooth-On, Dragon Skin 10 Medium) in \ac{3D}-printed moulds with pour and vent channels (Fig.~\ref{fig:phantom}a). These were then filled with an agar mixture with no acoustic scatterers to form solid inclusions which appear anechoic on B-mode \ac{US}. These objects were removed from the moulds, the sprues trimmed, and the objects were measured to obtain ground-truth dimensions (Fig.~\ref{fig:phantom}b). Four reference objects were created with varying geometric and symmetry properties: a sphere (Fig.~\ref{fig:phantom}c), an ellipsoid (spheroid) (Fig.~\ref{fig:phantom}d), a right circular cylinder (Fig.~\ref{fig:phantom}e), and an equilateral triangular prism (Fig.~\ref{fig:phantom}f). The inclusions were then set in a larger agar block, this time adding glass beads to introduce speckle noise in B-mode images (mimicking heterogeneous tissue), thereby creating contrast with the anechoic inclusions. A flat layer approximately 3 cm deep (one-third of the agar) was poured, allowed to cool, and mostly solidify, while the remainder was kept warm. The anechoic inclusions were then placed on top of the first layer, and a second layer was poured to surround and cover the inclusions (remaining two-thirds). Once fully set, the phantom was vacuum-sealed for longevity, and the appearance of inclusions was confirmed on B-mode. The recipe for making the agar mixtures is below and can be scaled to the required volume. STL files for \ac{3D}-printed parts are available at https://github.com/SMRTUS/US-QA-3D.

\noindent
For $\sim$250ml of agar:
\begin{itemize}
    \item 250~mL deionised water,
    \item 12.5~g agar-agar,
    \item 25~mL glycerol $\geq$99\% (increases speed of sound),
    \item 2.5~mL phenoxyethanol EHG (preservative),
    \item *10 g glass beads (soda lime glass microspheres, 3-6~$\mu$m diameter, introduces heterogeneity/speckle). 
\end{itemize}

\noindent
\begin{enumerate}
    \item Transfer deionised water into a Büchner flask and heat to 70–75~°C. 
    \item Slowly add agar-agar, stirring continuously to prevent clumping. Add the glycerol. *Optionally add the glass beads for a tissue-mimicking material, exclude for anechoic material. 
    \item Apply a vacuum and degass the mixture for 20–30~min. 
    \item Slowly release the vacuum, then carefully pour into the mould(s). Allow to cool and set at room temperature.
\end{enumerate}

\bibliographystyle{IEEEtran}
\bibliography{references}

@article{ng_early_2026,
	title = {Early {Adoption} of {Image}-{Guided} {Histotripsy} {Therapy} in {Interventional} {Oncology}: {Challenges} and {Opportunities} in the {UK}},
	issn = {0007-1285},
	shorttitle = {Early {Adoption} of {Image}-{Guided} {Histotripsy} {Therapy} in {Interventional} {Oncology}},
	url = {https://doi.org/10.1093/bjr/tqag047},
	doi = {10.1093/bjr/tqag047},
	urldate = {2026-03-18},
	journal = {British Journal of Radiology},
	author = {Ng, Helen Hoi Lam and Chan, Vinson Wai-Shun and Howell, Lewis and Shiwani, Taha and Zhong, Jim and Brandon, Jacqueline and Samson, Adel and Chandler, James and Mclaughlan, James and Wah, Tze Min},
	month = mar,
	year = {2026},
	pages = {tqag047},
}

@article{faoro_artificial_2023,
	title = {An {Artificial} {Intelligence}-{Aided} {Robotic} {Platform} for {Ultrasound}-{Guided} {Transcarotid} {Revascularization}},
	volume = {8},
	issn = {2377-3766},
	url = {https://ieeexplore.ieee.org/document/10057986},
	doi = {10.1109/LRA.2023.3251844},
	number = {4},
	urldate = {2023-12-03},
	journal = {IEEE Robotics and Automation Letters},
	author = {Faoro, Giovanni and Maglio, Sabina and Pane, Stefano and Iacovacci, Veronica and Menciassi, Arianna},
	month = apr,
	year = {2023},
	pages = {2349--2356},
}

@article{sun_automatic_2025,
	title = {Automatic {Robotic} {Ultrasound} for {3D} {Musculoskeletal} {Reconstruction}: {A} {Comprehensive} {Framework}},
	volume = {13},
	copyright = {http://creativecommons.org/licenses/by/3.0/},
	issn = {2227-7080},
	shorttitle = {Automatic {Robotic} {Ultrasound} for {3D} {Musculoskeletal} {Reconstruction}},
	url = {https://www.mdpi.com/2227-7080/13/2/70},
	doi = {10.3390/technologies13020070},
	language = {en},
	number = {2},
	urldate = {2026-03-13},
	journal = {Technologies},
	publisher = {Multidisciplinary Digital Publishing Institute},
	author = {Sun, Dezhi and Cappellari, Alessandro and Lan, Bangyu and Abayazid, Momen and Stramigioli, Stefano and Niu, Kenan},
	month = feb,
	year = {2025},
	pages = {70},
}

@article{schmidt_tracking_2024,
	title = {Tracking and mapping in medical computer vision: {A} review},
	volume = {94},
	issn = {1361-8423},
	shorttitle = {Tracking and mapping in medical computer vision},
	doi = {10.1016/j.media.2024.103131},
	language = {eng},
	journal = {Medical Image Analysis},
	author = {Schmidt, Adam and Mohareri, Omid and DiMaio, Simon and Yip, Michael C. and Salcudean, Septimiu E.},
	month = may,
	year = {2024},
	pages = {103131},
}

@article{chenhao_ultrasound_2026,
	title = {Ultrasound in medicine from 2014 to 2024: {A} bibliometric review},
	volume = {105},
	shorttitle = {Ultrasound in medicine from 2014 to 2024},
	url = {https://pmc.ncbi.nlm.nih.gov/articles/PMC12795102/},
	doi = {10.1097/MD.0000000000046890},
	language = {en},
	number = {2},
	urldate = {2026-02-24},
	journal = {Medicine},
	author = {Chenhao, Zhao and Ruqi, Zhou and Xiaozhuo, Sun and Jiabo, Wang and Jiao, Liu and Yajuan, Mu},
	month = jan,
	year = {2026},
	pages = {e46890},
}

@article{carbajal_improving_2013,
	title = {Improving {N}-wire phantom-based freehand ultrasound calibration},
	volume = {8},
	doi = {10.1007/s11548-013-0904-9},
	journal = {International journal of computer assisted radiology and surgery},
	author = {Carbajal, Guillermo and Lasso, Andras and Gomez, Alvaro and Fichtinger, Gabor},
	month = jul,
	year = {2013},
}

@article{tokuda_openigtlink_2009,
	title = {{OpenIGTLink}: an open network protocol for image-guided therapy environment},
	volume = {5},
	issn = {1478-5951},
	shorttitle = {{OpenIGTLink}},
	url = {https://www.ncbi.nlm.nih.gov/pmc/articles/PMC2811069/},
	doi = {10.1002/rcs.274},
	number = {4},
	urldate = {2023-02-21},
	journal = {The international journal of medical robotics and computer assisted surgery},
	author = {Tokuda, Junichi and Fischer, Gregory S. and Papademetris, Xenophon and Yaniv, Ziv and Ibanez, Luis and Cheng, Patrick and Liu, Haiying and Blevins, Jack and Arata, Jumpei and Golby, Alexandra J. and Kapur, Tina and Pieper, Steve and Burdette, Everette C. and Fichtinger, Gabor and Tempany, Clare M. and Hata, Nobuhiko},
	month = dec,
	year = {2009},
	pages = {423--434},
}

@misc{noauthor_moveitmoveit2_2025,
	title = {moveit/moveit2},
	copyright = {BSD-3-Clause},
	url = {https://github.com/moveit/moveit2},
	urldate = {2025-10-22},
	publisher = {MoveIt},
	month = oct,
	year = {2025},
}

@misc{tobias_frank_openigtlinkros2_igtl_bridge_2024,
	title = {openigtlink/ros2\_igtl\_bridge},
	copyright = {BSD-3-Clause},
	url = {https://github.com/openigtlink/ros2_igtl_bridge},
	urldate = {2025-10-22},
	publisher = {OpenIGTLink},
	author = {{Tobias Frank} and {Junichi Tokuda}},
	month = dec,
	year = {2024},
}

@misc{sun_alvinsunyixiaovrpn_mocap_2025,
	title = {alvinsunyixiao/vrpn\_mocap},
	copyright = {MIT},
	url = {https://github.com/alvinsunyixiao/vrpn_mocap},
	urldate = {2025-10-22},
	author = {Sun, Alvin},
	month = sep,
	year = {2025},
}

@article{li_multimodality_2018,
	title = {A multimodality imaging-compatible insertion robot with a respiratory motion calibration module designed for ablation of liver tumors: a preclinical study},
	volume = {34},
	issn = {0265-6736},
	shorttitle = {A multimodality imaging-compatible insertion robot with a respiratory motion calibration module designed for ablation of liver tumors},
	url = {https://doi.org/10.1080/02656736.2018.1456680},
	doi = {10.1080/02656736.2018.1456680},
	number = {8},
	urldate = {2026-01-12},
	journal = {International Journal of Hyperthermia},
	publisher = {Taylor \& Francis},
	author = {Li, Dongrui and Cheng, Zhigang and Chen, Gang and Liu, Fangyi and Wu, Wenbo and Yu, Jie and Gu, Ying and Liu, Fengyong and Ren, Chao and Liang, Ping},
	month = nov,
	year = {2018},
	pages = {1194--1201},
}

@book{russ_image_2018,
	address = {Boca Raton},
	edition = {7},
	title = {The {Image} {Processing} {Handbook}},
	isbn = {978-1-315-21411-5},
	doi = {10.1201/b18983},
	publisher = {CRC Press},
	author = {Russ, John C. and Neal, F. Brent},
	month = sep,
	year = {2018},
}

@article{otsu_threshold_1979,
	title = {A {Threshold} {Selection} {Method} from {Gray}-{Level} {Histograms}},
	volume = {9},
	issn = {2168-2909},
	url = {https://ieeexplore.ieee.org/document/4310076/},
	doi = {10.1109/TSMC.1979.4310076},
	number = {1},
	urldate = {2026-02-10},
	journal = {IEEE Transactions on Systems, Man, and Cybernetics},
	author = {Otsu, Nobuyuki},
	month = jan,
	year = {1979},
	pages = {62--66},
}

@article{priester_robotic_2013,
	title = {Robotic ultrasound systems in medicine},
	volume = {60},
	issn = {1525-8955},
	url = {https://ieeexplore.ieee.org/document/6470412},
	doi = {10.1109/TUFFC.2013.2593},
	number = {3},
	urldate = {2026-02-05},
	journal = {IEEE Transactions on Ultrasonics, Ferroelectrics, and Frequency Control},
	author = {Priester, Alan M. and Natarajan, Shyam and Culjat, Martin O.},
	month = mar,
	year = {2013},
	pages = {507--523},
}

@misc{northern_digital_inc_minimally_2024,
	title = {Minimally {Invasive} {Procedures} with {Electromagnetic} {Tracking} {Technology}},
	url = {https://www.ndigital.com/electromagnetic-tracking-technology/aurora/},
	language = {en-CA},
	urldate = {2026-01-14},
	journal = {NDI},
	author = {{Northern Digital Inc}},
	year = {2024},
}

@misc{universal_robots_universal_2025-1,
	title = {Universal {Robot} {PolyScope} {X} 10.11.1 {UR3e}},
	url = {https://www.universal-robots.com/manuals/EN/HTML/SW10_11/Content/Landingpages/WebPolyX/LandingPDFX.htm},
	urldate = {2025-10-22},
	author = {{Universal Robots}},
	year = {2025},
}

@article{peng_recent_2022,
	title = {Recent {Advances} in {Tracking} {Devices} for {Biomedical} {Ultrasound} {Imaging} {Applications}},
	volume = {13},
	issn = {2072-666X},
	url = {https://pmc.ncbi.nlm.nih.gov/articles/PMC9695235/},
	doi = {10.3390/mi13111855},
	number = {11},
	urldate = {2026-01-14},
	journal = {Micromachines},
	author = {Peng, Chang and Cai, Qianqian and Chen, Mengyue and Jiang, Xiaoning},
	month = oct,
	year = {2022},
	pages = {1855},
}

@article{iommi_evaluation_2020,
	title = {Evaluation of {3D} ultrasound for image guidance},
	volume = {15},
	issn = {1932-6203},
	url = {https://pmc.ncbi.nlm.nih.gov/articles/PMC7098612/},
	doi = {10.1371/journal.pone.0229441},
	number = {3},
	urldate = {2026-01-14},
	journal = {PLoS ONE},
	author = {Iommi, David and Hummel, Johann and Figl, Michael Lutz},
	month = mar,
	year = {2020},
	pages = {e0229441},
}

@misc{universal_robots_dh_2026,
	title = {{DH} {Parameters} for calculations of kinematics and dynamics},
	url = {https://www.universal-robots.com/articles/ur/application-installation/dh-parameters-for-calculations-of-kinematics-and-dynamics/},
	urldate = {2025-08-07},
	journal = {Universal Robots},
	author = {{Universal Robots}},
	month = nov,
	year = {2026},
}

@article{bekedam_intra-operative_2021,
	title = {Intra-operative resection margin model of tongue carcinoma using {3D} reconstructed ultrasound},
	volume = {4},
	issn = {2667-1476},
	url = {https://www.sciencedirect.com/science/article/pii/S2667147621001436},
	doi = {10.1016/j.adoms.2021.100154},
	urldate = {2026-01-12},
	journal = {Advances in Oral and Maxillofacial Surgery},
	author = {Bekedam, N. M. and Smit, J. N. and de Koekkoek - Doll, P. K. and van Alphen, M. J. A. and van Veen, R. L. P. and Karssemakers, L. H. E. and Karakullukcu, M. B. and Smeele, L. E.},
	month = oct,
	year = {2021},
	pages = {100154},
}

@inproceedings{chatelain_3d_2015,
	title = {{3D} ultrasound-guided robotic steering of a flexible needle via visual servoing},
	issn = {1050-4729},
	url = {https://ieeexplore.ieee.org/document/7139497},
	doi = {10.1109/ICRA.2015.7139497},
	urldate = {2026-01-12},
	booktitle = {2015 {IEEE} {International} {Conference} on {Robotics} and {Automation} ({ICRA})},
	author = {Chatelain, Pierre and Krupa, Alexandre and Navab, Nassir},
	month = may,
	year = {2015},
	pages = {2250--2255},
}

@article{wang_task_2022,
	title = {Task model-specific operator skill assessment in routine fetal ultrasound scanning},
	volume = {17},
	issn = {1861-6410},
	url = {https://pmc.ncbi.nlm.nih.gov/articles/PMC9307537/},
	doi = {10.1007/s11548-022-02642-y},
	number = {8},
	urldate = {2026-01-12},
	journal = {International Journal of Computer Assisted Radiology and Surgery},
	author = {Wang, Yipei and Yang, Qianye and Drukker, Lior and Papageorghiou, Aris and Hu, Yipeng and Noble, J. Alison},
	year = {2022},
	pages = {1437--1444},
}

@article{huang_review_2017,
	title = {A {Review} on {Real}-{Time} {3D} {Ultrasound} {Imaging} {Technology}},
	volume = {2017},
	issn = {2314-6141},
	doi = {10.1155/2017/6027029},
	language = {eng},
	journal = {BioMed Research International},
	author = {Huang, Qinghua and Zeng, Zhaozheng},
	year = {2017},
	pages = {6027029},
}

@article{almuhanna_carotid_2015,
	title = {Carotid plaque morphometric assessment with three-dimensional ultrasound imaging},
	volume = {61},
	issn = {0741-5214},
	url = {https://www.sciencedirect.com/science/article/pii/S0741521414018667},
	doi = {10.1016/j.jvs.2014.10.003},
	number = {3},
	urldate = {2025-12-10},
	journal = {Journal of Vascular Surgery},
	author = {AlMuhanna, Khalid and Hossain, Md Murad and Zhao, Limin and Fischell, Jonathan and Kowalewski, Gregory and Dux, Moira and Sikdar, Siddhartha and Lal, Brajesh K.},
	month = mar,
	year = {2015},
	pages = {690--697},
}

@article{chung_freehand_2017,
	title = {Freehand three-dimensional ultrasound imaging of carotid artery using motion tracking technology},
	volume = {74},
	issn = {0041-624X},
	url = {https://www.sciencedirect.com/science/article/pii/S0041624X16302001},
	doi = {10.1016/j.ultras.2016.09.020},
	urldate = {2025-11-19},
	journal = {Ultrasonics},
	author = {Chung, Shao-Wen and Shih, Cho-Chiang and Huang, Chih-Chung},
	month = feb,
	year = {2017},
	pages = {11--20},
}

@article{howell_histotripsy_2025,
	title = {Histotripsy: {New} {Frontiers} for {Noninvasive} {Ablation}},
	volume = {9},
	issn = {2542-7075},
	shorttitle = {Histotripsy},
	url = {https://journals.lww.com/ajir/fulltext/2025/07000/histotripsy__new_frontiers_for_noninvasive.1.aspx},
	doi = {10.4103/AJIR-2025-19},
	language = {en-US},
	number = {2},
	urldate = {2025-12-09},
	journal = {The Arab Journal of Interventional Radiology},
	author = {Howell, Lewis J. and Chan, Vinson Wai-Shun and Zhong, Jim and Ng, Helen Hoi-Lam and Chandler, James H. and McLaughlan, James R. and Wah, Tze Min},
	month = dec,
	year = {2025},
	pages = {61},
}

@article{yandong_establishment_2024,
	title = {Establishment and preliminary application of personalized three‐dimensional reconstruction of thyroid gland with automatic detection of thyroid nodules based on ultrasound videos},
	volume = {25},
	issn = {1526-9914},
	url = {https://pmc.ncbi.nlm.nih.gov/articles/PMC11163481/},
	doi = {10.1002/acm2.14332},
	number = {6},
	urldate = {2025-12-09},
	journal = {Journal of Applied Clinical Medical Physics},
	author = {Yandong, Huang and Shiqi, Zhang and Lanting, Jia and Wenxin, Hu and Leyao, Chen and Hejing, Huang},
	month = mar,
	year = {2024},
	pages = {e14332},
}

@article{zielke_rsv_2022,
	title = {{RSV}: {Robotic} {Sonography} for {Thyroid} {Volumetry}},
	volume = {7},
	issn = {2377-3766},
	shorttitle = {{RSV}},
	url = {https://ieeexplore.ieee.org/document/9695372},
	doi = {10.1109/LRA.2022.3146542},
	number = {2},
	urldate = {2025-12-03},
	journal = {IEEE Robotics and Automation Letters},
	author = {Zielke, John and Eilers, Christine and Busam, Benjamin and Weber, Wolfgang and Navab, Nassir and Wendler, Thomas},
	month = apr,
	year = {2022},
	pages = {3342--3348},
}

@article{merouche_robotic_2016,
	title = {A {Robotic} {Ultrasound} {Scanner} for {Automatic} {Vessel} {Tracking} and {Three}-{Dimensional} {Reconstruction} of {B}-{Mode} {Images}},
	volume = {63},
	issn = {1525-8955},
	url = {https://ieeexplore.ieee.org/document/7322288},
	doi = {10.1109/TUFFC.2015.2499084},
	number = {1},
	urldate = {2025-11-19},
	journal = {IEEE Transactions on Ultrasonics, Ferroelectrics, and Frequency Control},
	author = {Merouche, Samir and Allard, Louise and Montagnon, Emmanuel and Soulez, Gilles and Bigras, Pascal and Cloutier, Guy},
	month = jan,
	year = {2016},
	pages = {35--46},
}

@article{poulin_interference_2002,
	title = {Interference during the use of an electromagnetic tracking system under {OR} conditions},
	volume = {35},
	issn = {0021-9290},
	url = {https://www.sciencedirect.com/science/article/pii/S0021929002000362},
	doi = {10.1016/S0021-9290(02)00036-2},
	number = {6},
	urldate = {2024-11-11},
	journal = {Journal of Biomechanics},
	author = {Poulin, François and Amiot, L. -P.},
	month = jun,
	year = {2002},
	pages = {733--737},
}

@article{aurand_accuracy_2017,
	title = {Accuracy map of an optical motion capture system with 42 or 21 cameras in a large measurement volume},
	volume = {58},
	issn = {0021-9290},
	url = {https://www.sciencedirect.com/science/article/pii/S0021929017302580},
	doi = {10.1016/j.jbiomech.2017.05.006},
	urldate = {2025-11-12},
	journal = {Journal of Biomechanics},
	author = {Aurand, Alexander M. and Dufour, Jonathan S. and Marras, William S.},
	month = jun,
	year = {2017},
	pages = {237--240},
}

@misc{optitrack_naturalpoint_inc_calibration_2022,
	title = {Calibration v2.3 {OptiTrack} {Documentation}},
	url = {https://docs.optitrack.com/v2.3/motive/calibration},
	language = {en},
	urldate = {2025-11-12},
	author = {{OptiTrack (NaturalPoint Inc.)}},
	month = oct,
	year = {2022},
}

@misc{computerized_imaging_reference_systems_inc_multi-purpose_2013,
	title = {Multi-{Purpose}, {Multi}-{Tissue} {Ultrasound} {Phantom}},
	author = {{Computerized Imaging Reference Systems Inc.}},
	year = {2013},
}

@article{zhou_deep_2025,
	title = {A deep learning based ultrasound diagnostic tool driven by {3D} visualization of thyroid nodules},
	volume = {8},
	copyright = {2025 The Author(s)},
	issn = {2398-6352},
	url = {https://www.nature.com/articles/s41746-025-01455-y},
	doi = {10.1038/s41746-025-01455-y},
	language = {en},
	number = {1},
	urldate = {2025-11-11},
	journal = {npj Digital Medicine},
	publisher = {Nature Publishing Group},
	author = {Zhou, Yahan and Chen, Chen and Yao, Jincao and Yu, Jiabin and Feng, Bojian and Sui, Lin and Yan, Yuqi and Chen, Xiayi and Liu, Yuanzhen and Zhang, Xiao and Wang, Hui and Pan, Qianmeng and Zou, Weijie and Zhang, Qi and Lin, Lu and Xu, Chenke and Yuan, Shengxing and He, Qingquan and Ding, Xiaofan and Liang, Ping and Wang, Vicky Yang and Xu, Dong},
	month = feb,
	year = {2025},
	pages = {126},
}

@article{mozaffari_freehand_2017,
	title = {Freehand 3-{D} {Ultrasound} {Imaging}: {A} {Systematic} {Review}},
	volume = {43},
	issn = {1879-291X},
	shorttitle = {Freehand 3-{D} {Ultrasound} {Imaging}},
	doi = {10.1016/j.ultrasmedbio.2017.06.009},
	language = {eng},
	number = {10},
	journal = {Ultrasound in Medicine \& Biology},
	author = {Mozaffari, Mohammad Hamed and Lee, Won-Sook},
	month = oct,
	year = {2017},
	pages = {2099--2124},
}

@article{bozorgi_gpu-based_2015,
	title = {{GPU}-based multi-volume ray casting within {VTK} for medical applications},
	volume = {10},
	issn = {1861-6429},
	doi = {10.1007/s11548-014-1069-x},
	language = {eng},
	number = {3},
	journal = {International Journal of Computer Assisted Radiology and Surgery},
	author = {Bozorgi, Mohammadmehdi and Lindseth, Frank},
	month = mar,
	year = {2015},
	pages = {293--300},
}

@incollection{mohamed_survey_2019,
	title = {A {Survey} on {3D} {Ultrasound} {Reconstruction} {Techniques}},
	isbn = {978-1-78984-018-6},
	url = {https://www.intechopen.com/chapters/63949},
	doi = {10.5772/intechopen.81628},
	language = {en},
	urldate = {2025-10-28},
	booktitle = {Artificial {Intelligence} - {Applications} in {Medicine} and {Biology}},
	publisher = {IntechOpen},
	author = {Mohamed, Farhan and Siang, Chan Vei and Mohamed, Farhan and Siang, Chan Vei},
	month = apr,
	year = {2019},
}

@article{kronke_tracked_2022,
	title = {Tracked {3D} ultrasound and deep neural network-based thyroid segmentation reduce interobserver variability in thyroid volumetry},
	volume = {17},
	issn = {1932-6203},
	url = {https://journals.plos.org/plosone/article?id=10.1371/journal.pone.0268550},
	doi = {10.1371/journal.pone.0268550},
	language = {en},
	number = {7},
	urldate = {2022-08-23},
	journal = {PLOS ONE},
	publisher = {Public Library of Science},
	author = {Krönke, Markus and Eilers, Christine and Dimova, Desislava and Köhler, Melanie and Buschner, Gabriel and Schweiger, Lilit and Konstantinidou, Lemonia and Makowski, Marcus and Nagarajah, James and Navab, Nassir and Weber, Wolfgang and Wendler, Thomas},
	month = jul,
	year = {2022},
	pages = {e0268550},
}

@misc{perk_lab_pluslib_2025,
	title = {{PlusLib}: {NDI} {Vega}, {Polaris} and {Aurora} pose trackers},
	url = {http://perk-software.cs.queensu.ca/plus/doc/nightly/dev/DeviceNDI.html},
	urldate = {2025-10-28},
	author = {{Perk Lab}},
	month = aug,
	year = {2025},
}

@article{daunizeau_robot-assisted_2020,
	title = {Robot-assisted ultrasound navigation platform for {3D} {HIFU} treatment planning: {Initial} evaluation for conformal interstitial ablation},
	volume = {124},
	issn = {0010-4825},
	shorttitle = {Robot-assisted ultrasound navigation platform for {3D} {HIFU} treatment planning},
	url = {https://www.sciencedirect.com/science/article/pii/S0010482520302766},
	doi = {10.1016/j.compbiomed.2020.103941},
	urldate = {2025-07-25},
	journal = {Computers in Biology and Medicine},
	author = {Daunizeau, L. and Nguyen, A. and Le Garrec, M. and Chapelon, J. Y. and N'Djin, W. A.},
	month = sep,
	year = {2020},
	pages = {103941},
}

@article{connolly_bridging_2022,
	title = {Bridging {3D} {Slicer} and {ROS2} for {Image}-{Guided} {Robotic} {Interventions}},
	volume = {22},
	issn = {1424-8220},
	url = {https://www.ncbi.nlm.nih.gov/pmc/articles/PMC9324680/},
	doi = {10.3390/s22145336},
	number = {14},
	urldate = {2023-02-21},
	journal = {Sensors (Basel, Switzerland)},
	author = {Connolly, Laura and Deguet, Anton and Leonard, Simon and Tokuda, Junichi and Ungi, Tamas and Krieger, Axel and Kazanzides, Peter and Mousavi, Parvin and Fichtinger, Gabor and Taylor, Russell H.},
	month = jul,
	year = {2022},
	pages = {5336},
}

@article{ungi_open-source_2016,
	series = {20th anniversary of the {Medical} {Image} {Analysis} journal ({MedIA})},
	title = {Open-source platforms for navigated image-guided interventions},
	volume = {33},
	issn = {1361-8415},
	url = {https://www.sciencedirect.com/science/article/pii/S1361841516300895},
	doi = {10.1016/j.media.2016.06.011},
	language = {en},
	urldate = {2023-02-21},
	journal = {Medical Image Analysis},
	author = {Ungi, Tamas and Lasso, Andras and Fichtinger, Gabor},
	month = oct,
	year = {2016},
	pages = {181--186},
}

@inproceedings{wein_three-dimensional_2020,
	address = {Cham},
	series = {Lecture {Notes} in {Computer} {Science}},
	title = {Three-{Dimensional} {Thyroid} {Assessment} from {Untracked} {2D} {Ultrasound} {Clips}},
	isbn = {978-3-030-59716-0},
	doi = {10.1007/978-3-030-59716-0_49},
	language = {en},
	booktitle = {Medical {Image} {Computing} and {Computer} {Assisted} {Intervention} – {MICCAI} 2020},
	publisher = {Springer International Publishing},
	author = {Wein, Wolfgang and Lupetti, Mattia and Zettinig, Oliver and Jagoda, Simon and Salehi, Mehrdad and Markova, Viktoria and Zonoobi, Dornoosh and Prevost, Raphael},
	editor = {Martel, Anne L. and Abolmaesumi, Purang and Stoyanov, Danail and Mateus, Diana and Zuluaga, Maria A. and Zhou, S. Kevin and Racoceanu, Daniel and Joskowicz, Leo},
	year = {2020},
	pages = {514--523},
}

@article{boers_3d_2024,
	title = {{3D} ultrasound guidance for radiofrequency ablation in an anthropomorphic thyroid nodule phantom},
	volume = {8},
	issn = {2509-9280},
	url = {https://doi.org/10.1186/s41747-024-00513-6},
	doi = {10.1186/s41747-024-00513-6},
	language = {en},
	number = {1},
	urldate = {2025-05-06},
	journal = {European Radiology Experimental},
	author = {Boers, Tim and Braak, Sicco J. and Brink, Wyger M. and Versluis, Michel and Manohar, Srirang},
	month = oct,
	year = {2024},
	pages = {115},
}

@article{an_ultrasound_2017,
	title = {An {Ultrasound} {Imaging}-{Guided} {Robotic} {HIFU} {Ablation} {Experimental} {System} and {Accuracy} {Evaluations}},
	volume = {2017},
	issn = {1176-2322},
	url = {https://www.hindawi.com/journals/abb/2017/5868695/},
	doi = {10.1155/2017/5868695},
	language = {en},
	urldate = {2023-02-14},
	journal = {Applied Bionics and Biomechanics},
	publisher = {Hindawi},
	author = {An, Chih Yu and Syu, Jia Hao and Tseng, Ching Shiow and Chang, Chih-Ju},
	month = apr,
	year = {2017},
	pages = {e5868695},
}

@article{lasso_plus_2014,
	title = {{PLUS}: {Open}-source toolkit for ultrasound-guided intervention systems},
	volume = {61},
	issn = {0018-9294},
	shorttitle = {{PLUS}},
	doi = {10.1109/TBME.2014.2322864},
	journal = {IEEE Transactions on Biomedical Engineering},
	author = {Lasso, Andras and Heffter, Tamas and Rankin, Adam and Pinter, Csaba and Ungi, Tamas and Fichtinger, Gabor},
	year = {2014},
	pages = {2527--2537},
}

\end{document}